\def\year{2021}\relax
\documentclass[letterpaper]{article} 
\usepackage{amsfonts}
\usepackage{amsmath}
\usepackage{mathrsfs}
\usepackage{aaai21}  
\usepackage{times}  
\usepackage{helvet} 
\usepackage{courier}  
\usepackage[hyphens]{url}  
\usepackage{graphicx} 
\usepackage{natbib}  
\usepackage{caption} 
\usepackage{color}
\usepackage{amsmath}
\usepackage{extarrows}
\newcommand{\argmin}[1]{\underset{#1}{\operatorname{arg}\,\operatorname{min}}\;}
\usepackage{graphicx}
\usepackage{subfigure}
\usepackage{algorithm}  
\usepackage[noend]{algpseudocode}
\title{Uncertainty-aware INVASE: Enhanced Breast Cancer Diagnosis Feature Selection}
\author {
    Jia-Xing Zhong,\textsuperscript{\rm 1} 
    Hongbo Zhang \textsuperscript{\rm 2}\\
}

\affiliations {
    \textsuperscript{\rm 1} School of Electronic and Computer Engineering, Peking University \\
    \textsuperscript{\rm 2} Department of Computer and Information Sciences, Virginia Military Institute \\
    jxzhong@pku.edu.cn, hbzhang@vt.edu
}

\begin{document}

\maketitle

\begin{abstract}
In this paper, we present an uncertainty-aware INVASE to quantify predictive confidence of healthcare  problem. By introducing learnable Gaussian distributions, we leverage their variances to measure the degree of uncertainty. Based on the vanilla INVASE, two additional modules are proposed, \text{i.e.}, an uncertainty quantification module in the predictor, and a reward shaping module in the selector. We conduct extensive experiments on UCI-WDBC dataset. Notably, our method eliminates almost all predictive bias with only about 20\% queries, while the uncertainty-agnostic counterpart requires nearly 100\% queries. 
The open-source implementation with a detailed tutorial is available at \url{https://github.com/jx-zhong-for-academic-purpose/Uncertainty-aware-INVASE/blob/main/tutorial_invase%2B.ipynb}.
\end{abstract}

\section{Introduction}

Breast cancer is an increasing health problem ~\cite{howell2014risk}. One in Eight U.S. women will develop invasive breast cancer in her life time. Early diagnosis of Breast cancer is important. Among them, conventional global feature based machine learning method has only achieved limited successes ~\cite{wang2016deep}. High dimension instance-wise feature selection is an emerging machine learning approach, on which the relevant subset of features should be discovered \emph{for each individual data sample}. To address that problem, researchers have proposed several valuable models~\cite{shrikumar2017learning,yoon2018invase,chen2018learning,lundberg2017unified}. Among them, learning to explain~\cite{chen2018learning} has built the foundation of instance feature selection and explanation of the features as well using a mutual information model.  As one of the state-of-the-art algorithms, INVASE~\cite{yoon2018invase} further extends the learning to explain using a baseline network and a predictor to train a selector in the actor-critic manner, which allows variable-size feature selection. 

Existing instance-wise feature selectors are devised to achieve high performance in target tasks. However, they ignore another important goal: \emph{capture the confidence of their outputs}. The lack of accurate confidence interval will lead to deviated estimate. {In practice, that may lead to over-confident yet incorrect predictions, which is likely dangerous particularly in the application scenario of healthcare~\cite{tonekaboni2019clinicians} such as false positive and false negative.} Since medical services are extremely complex and hardly fault-tolerance, a confidence-agnostic algorithm is undesirable. Thus, an uncertainty-aware approach to mitigate the over confidence problem should be introduced to instance-wise feature selection, for the purpose of \emph{avoiding potential error decisions}.

In this paper, we enhance INVASE to \emph{quantify its predictive confidence by learning an uncertainty estimation}. The vanilla INVASE optimizes the predictor by treating data points as samples from a set of distributions with Dirac delta probability density functions, whereas our model regards data as samples from learnable uncertainty-aware distributions. To be specific, we establish our model with series of Gaussian distributions, of which the corresponding \emph{variances measure the degree of uncertainty}. Our model is completely consistent with the vanilla INVASE in an extreme condition, \textit{i.e.}, the model ``thinks'' that every prediction is absolutely certain. In our work, two modules are added to the vanilla INVASE, \emph{viz.}, uncertainty quantification and reward shaping. The former estimates the uncertainty of selected features for the predictor, while the latter assists in improving such estimation via the selector. 

To demonstrate the efficacy of our presented extension for INVASE, we conduct extensive experiments on a real-world medical dataset \textit{Wisconsin Diagnostic Breast Cancer} (UCI-WDBC)~\cite{Dua:2019}. Experimental results show the superiority of our certainty-aware approach: at only about $20\%$ query rates, our model correct almost all predictive bias. To achieve an equal performance gain, certainty-agnostic counterparts require to query about nearly $100\%$ testing data. 

In summary, the contribution of this paper is three-fold:
\begin{itemize}
\item Based on INVASE, we put forward an uncertainty-aware extension. To the best of our knowledge, our model is the first instance-wise feature selector to quantify predictive uncertainty. That is beneficial to discover potential mistakes of the model output.
\item Theoretically, the vanilla version of INVASE can be considered as a particular case of ours. As a seamlessly backward-compatible extension, our implementation only needs two modules: uncertainty quantification for the predictor and reward shaping for the selector. 
\item Experimentally, we evaluate our model on the UCI-WDBC dataset from two aspects, \textit{i.e.}, overall performance of the intact model, along with in-depth studies of every component module. {For reproducible research, the source codes are provided online. }
\end{itemize}
\section{Related Work}
\textbf{Instance-wise feature selectors} have only been studied recently~\cite{shrikumar2017learning,yoon2018invase,chen2018learning,lundberg2017unified}, unlike the well-developed global ones~\cite{LIN201592,candes2018panning,lin2012maximal}. Different from prior work, our uncertainty-aware INVASE is able to quantify the confidence for instance-wise predictions.


\textbf{Uncertainty quantification} has two main categories from the perspective of Bayesian modeling~\cite{u1,u2,u3,u4}, \textit{i.e.}, \emph{aleatoric} and \emph{epistemic} uncertainty. Aleatoric uncertainty (a.k.a. data uncertainty) originates from the information bias of datasets, \textit{e.g.}, noisy observations in the data. Epistemic uncertainty (a.k.a. model uncertainty) stems from the unseen inputs of a model, \textit{e.g.}, insufficient training samples. Under such a taxonomy, the uncertainty studied in this paper can be viewed as a type of aleatoric uncertainty. 
\section{Methodology}
\begin{figure}[h!]
  \centering
  \includegraphics[width=0.46\textwidth]{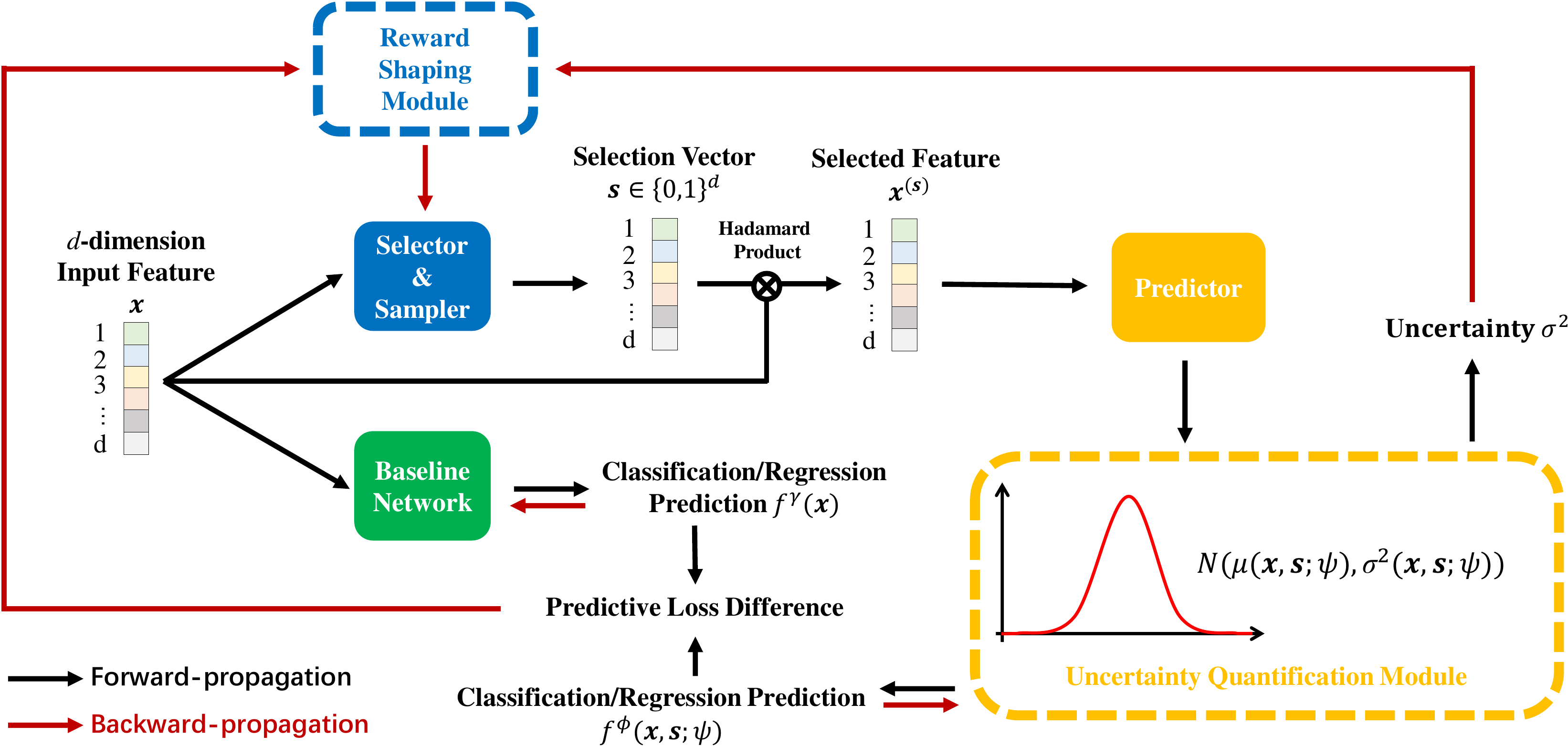}
\caption{\emph{Framework of uncertainty-aware INVASE.} Our baseline network is exactly the same as the vanilla INVASE, whereas we make changes in the selector and the predictor. The two new modules over the vanilla INVASE are denoted with \emph{dotted boxes}. Based on the predictor, we apply an uncertainty quantification module to estimate the predictive uncertainty for selected features. Based on the selector, we use a reward shaping module to guide the process of uncertainty exploration toward a more precise result.}
\label{fig:framewrok}
\end{figure}
\subsection{Problem Statement}
Given a continuous label space $\mathcal{Y}=\mathbb{R}$ or a proper subset of $\mathbb{R}$  (or its discrete \textit{c}-class counterpart $\mathcal{Y}=\{1,2,..,c\}$), we denote a label as $Y \in \mathcal{Y}$. $\mathcal{X}=\mathcal{X}_1 \times \mathcal{X}_2 \times ...\times \mathcal{X}_d$ represents a d-dimension input feature space, and let $X=(X_1,X_2,...,X_d) \in \mathcal{X}$ be a random variable. Following the notations of INVASE, a selection vector ${\bf s}=\{0,1\}^d$ indicates that the $i^{th}$-dimension variable is selected if ${\bf s}_i=1$, otherwise it is not selected.
In the formulation of \emph{instance-wise feature selection}, we are required to obtain an optimal selection $\bf s$ for a certain realization ${\bf x} \in \mathcal{X}$ of $X$. The corresponding suppressed feature vector for the $i^{th}$ dimension is defined as:
\begin{equation}
{\bf x}_i^{({\bf s})}=\left\{
\begin{array}{ccc}
{{\bf x}_i} & & ({\bf s}_i=1)\\
{*} & & ({\bf s}_i=0)\\
\end{array}\right.  \,,
\end{equation}
where $*$ refers to that the $i^{th}$ dimension is not chosen. We define a selection function $\mathcal{S}:\mathcal{X} \mapsto \{0,1\}^d$ for the $d$-dimension instance-wise feature:  
\begin{equation}\label{dist}
(Y|X^{S({\bf x})}={\bf x}^{S({\bf x})})\xlongequal[]{dis.}(Y|X={\bf x})  \,,
\end{equation}
where $\xlongequal[]{dis.}$ means distributional equality and $S({\bf x})$ is minimal in accordance with the equality.
\subsection{Recapitulation of the Vanilla INVASE}
In the vanilla INVASE, Kullback-Leibler (KL) divergence is leveraged to measure the ``difference'' between the two distributions in Equation~(\ref{dist}). To minimize the KL divergence, INVASE defines a loss estimator $\hat{l}({\bf x}, {\bf s})$ to approximate it for regression problems w.r.t. the training dataset $D$:
\begin{equation}\label{raw_invase}
\hat{l}({\bf x}, {\bf s})=-\sum_{({\bf x},y)\in D}(||y-f^\phi({\bf x},{\bf s})||_2-||y-f^\gamma({\bf x})||_2) \,,
\end{equation}
where $f^\gamma$ and $f^\phi$ is the {baseline network fed with the whole feature set $\bf x$ and the predictor relied on the selected feature subset $({\bf x},{\bf s})$} parameterized by $\gamma$ and $\phi$, respectively; $||\cdot||_2$ refers to the value of $\mathscr{l_2}$ loss. Intuitively, the INVASE is intended to \emph{choose a subset $({\bf x},{\bf s})$, upon which the performance surpasses that based on all features $\bf x$ as much as possible}. As for classification problems, \cite{yoon2018invase} point out that optimizing the $\mathscr{l_2}$ loss is equivalent to minimizing the KL divergence for classification when the distribution of $Y|X$ is Gaussian. Therefore, we only discuss the model under the regression setting in our remaining paper for simplicity. 
\subsection{Extra Optimizing Objective beyond INVASE}
In Equation~(\ref{raw_invase}), the term $||y-f^\phi({\bf x},{\bf s})||_2$ is designed to estimate the predictive loss of selected features ${\bf x}^{({\bf s})}$. By doing this, it treats an observation $({\bf x}, y)$ in the dataset $D$ as a sample from a distribution of which the probability density function is a \emph{Dirac delta function $\delta$ without capturing uncertainty}:
\begin{equation}
P_D(y|{\bf x},{\bf s})=P^\delta(y-f^\phi({\bf x},{\bf s}))\,.
\end{equation}
Unlike INVASE, our model regards $({\bf x}, y)$ in the training dataset $D$ as a sample from a \emph{learnable uncertainty-aware distribution parameterized by $\psi$}:
\begin{equation}\label{eq:p_psi}
P_D(y|{\bf x},{\bf s})=P^\psi(y-f^\phi({\bf x},{\bf s}))\,.
\end{equation}
In this context, our modeling parameters $\psi$ can be trained by minimizing its KL Divergence from the dataset's distribution:
\begin{equation}\label{kl_obj}
\psi^*= \argmin{\psi} E_{{\bf x} \sim P_D({\bf x})}( d_{KL}(P_D(y|{\bf x},{\bf s})||P^\psi(y-f^\phi({\bf x},{\bf s}))))\,,
\end{equation}
where $d_{KL}$ is the KL divergence and $E$ is the mathematical expectation. Besides the original goals of INVASE, our \emph{extra optimizing objective} is to obtain $\psi^*$ with a suitable distribution type instead of simply using the Dirac delta function. Conceptually, \emph{our framework is highly scalable since any distribution with differentiable parameters is an eligible tool}.
\subsection{Uncertain-aware INVASE}
Following some prior research~\cite{u1,u5} on modeling uncertainty, we specify a set of Gaussian distributions to analyze the uncertainty of instance-wise feature selection.

As illustrated in Figure~\ref{fig:framewrok}, three neural networks constitute our whole model. Among them, the baseline network $f^\gamma$ is identical with its counterpart in vanilla INVASE. As for the other two networks, we introduce two additional modules respectively, uncertainty quantification of the predictor and reward shaping of the selector networks.
\subsubsection{Predictor with Uncertainty Quantification}
Fed into the selected feature, the predictor outputs the corresponding regression result to evaluate the performance of selection. We devise an uncertainty quantification module to capture the output uncertainty of our predictor. Specifically, we introduce a network branch parameterized by $\psi$ to learn the mean value $\mu ({\bf x},{\bf s};\psi)$ and variance $\sigma^2 ({\bf x},{\bf s};\psi)$ for a certain selected feature ${\bf x}^{({\bf s})}$. To optimize $\psi$ as described in Equation~(\ref{kl_obj}), we minimize the negative log-likelihood cost in the predictor $\phi$:
\begin{equation}
\begin{aligned}
l^\phi({\bf x},{\bf s};\psi)&=-\log P_D^\phi(y|{\bf x},{\bf s};\psi)\\&= \frac{\log \sigma^2 ({\bf x},{\bf s};\psi)}{2}+\frac{||y-\mu ({\bf x},{\bf s};\psi)||_2}{2\sigma^2 ({\bf x},{\bf s};\psi)}+constant\,.
\end{aligned}
\label{log_l}
\end{equation}
Intuitively, if $\mu ({\bf x},{\bf s};\psi)$ easily fits $y$ (with low uncertainty), the predictive bias term $||y-\mu ({\bf x},{\bf s};\psi)||_2$ tends to be small so that the first term $\frac{\log \sigma^{2} ({\bf x},{\bf s};\psi)}{2}$ dominates our cost. By minimizing the cost in this case, we will obtain a smaller variance $\sigma^2$. Otherwise, $\sigma^2$ is inclined to be larger if the label $y$ is difficult to approximate (with high uncertainty). 

Thus, we \textit{quantify uncertainty with $\sigma^2$: a larger variance means higher uncertainty}. Based on the sample-wise Gaussian distribution $N(\mu,\sigma^2)$, our loss estimator is computed as:
\begin{equation}
\begin{split}
\hat{l}({\bf x},{\bf s})=-\sum_{({\bf x},y)\in D}((\frac{\log \sigma^2 ({\bf x},{\bf s};\psi)}{2}+\frac{||y-\mu ({\bf x},{\bf s};\psi)||_2}{2\sigma^2 ({\bf x},{\bf s};\psi)})\\-||y-f^\gamma({\bf x})||_2)\,,
\end{split}
\end{equation}
where the meaning of all notations follows Equation~(\ref{raw_invase}) and (\ref{log_l}). In practice, we append a fully-connected branch to the predictor for computation of $\psi$.

\subsubsection{Selector with Reward Shaping}
The selector $f^\theta$ is trained to choose an appropriate subset of instance-wise variables. The reward of its original version is defined as:
\begin{equation}\label{ori_Reward}
R({\bf x},{\bf s})={-}{\hat{l}({\bf x},{\bf s})}-\lambda||{\bf s}||_0\,,
\end{equation}
where the $\mathscr{l_0}$-norm $||{\bf s}||_0$ constrains the dimension number of selected features and $\lambda$ is a weighting hyper-parameter. In our model, a reward shaping module encourages the selector to explore more uncertain samples, which assists in estimating uncertainty more accurately. We shape the reward by adding an uncertainty preference term to optimize the policy of our selector $\theta$:
\begin{equation}\label{RS}
R({\bf x},{\bf s})=\omega \sigma^2({\bf x},{\bf s};\psi){-}{\hat{l}({\bf x},{\bf s})}-\lambda||{\bf s}||_0\,,
\end{equation}
where $\omega$ is a hyper-parameter to control the balance between uncertainty preference and the other rewards. As shown in Section~\ref{sec:ex}, the reward shaping module makes the uncertainty adequately explored. In the testing phase, the prediction $f^\phi({\bf x},{\bf s};\psi)$ for an input feature $\bf x$ with selection vector $\bf s$ gives the regression result as $\mu ({\bf x},{\bf s};\psi)$ and the uncertainty as $\sigma^2 ({\bf x},{\bf s};\psi)$. In {\bf Appendix}, we will discuss the relationship between our model and the vanilla INVASE. 

\section{Experiments}
\paragraph{Criteria for Evaluation}Suppose that we are utilizing the uncertainty-aware INVASE to diagnose breast cancers with selected features. When we meet a highly uncertain prediction of our model, we will naturally query the exact answer from a skillful doctor. According to uncertainty scores, our goal is to achieve higher performance with fewer queries. Thus, we evaluate the model by observing the \emph{performance gain on test data across different query rates}. For simplicity, we assume that the doctor does not make any error, \emph{i.e.}, answers to all queries are always right. The queried samples by doctor consists of the data with uncertain of prediction thus need to be verified by doctor for correction. 
\paragraph{Dataset and Evaluation Metric}As the given implementation of INVASE, we carry out experiments on UCI-WDBC~\cite{Dua:2019} dataset, which has 569 records of a breast cancer diagnosis with $30$-dimension features. Following the original setting, we hold out 80\% data for training and randomly sample the test set $20$ times. The weighting hyper-parameter of reward shaping $\omega$ is set as $0.1$ empirically. In all the experiments, we keep the default
settings identical to the vanilla INVASE if not specified particularly. We keep the same performance metrics as the vanilla INVASE, mainly area under the curve of receiver operating characteristic (AUC-ROC) and average precision (a.k.a. area under the curve of Precision-Recall, AUC-PR). 
\paragraph{Benchmarks for Comparison}Since no prior instance-wise feature selector explicitly quantifies the predictive uncertainty, there does not exist prior work for comparison. Hereby, we introduce two benchmarks, \emph{i.e.}, ``Oracle'' and ``w/o Uncertainty''. \textit{Oracle} is an ideal selection strategy: a sample with the largest predictive bias from the ground truth takes precedence. That can be deemed as the upper bound of an uncertainty-aware model since every query is capable of maximizing the performance gain. Here the query means the data that are in need of further doctor verification.  Another selection method denoted as \textit{w/o Uncertainty} is that we know nothing about uncertainty and randomly choose our queries, corresponding to uncertainty-agnostic models (\textit{e.g.}, the vanilla INVASE). The uncertainty here refers to the probability of a sample needs to be sent to doctor for further verification. As for the \textit{our model}, we just prioritize the query about an uncertain prediction: queries are submitted in descending order of uncertainty scores (\textit{i.e.}, the variance $\sigma^2$).
\begin{figure}[h!]
\centering
\subfigure[Predictive Bias from Ground Truths]{\label{fig:bias}
\begin{minipage}[t]{0.15\textwidth}
\centering
\includegraphics[width=.965\textwidth]{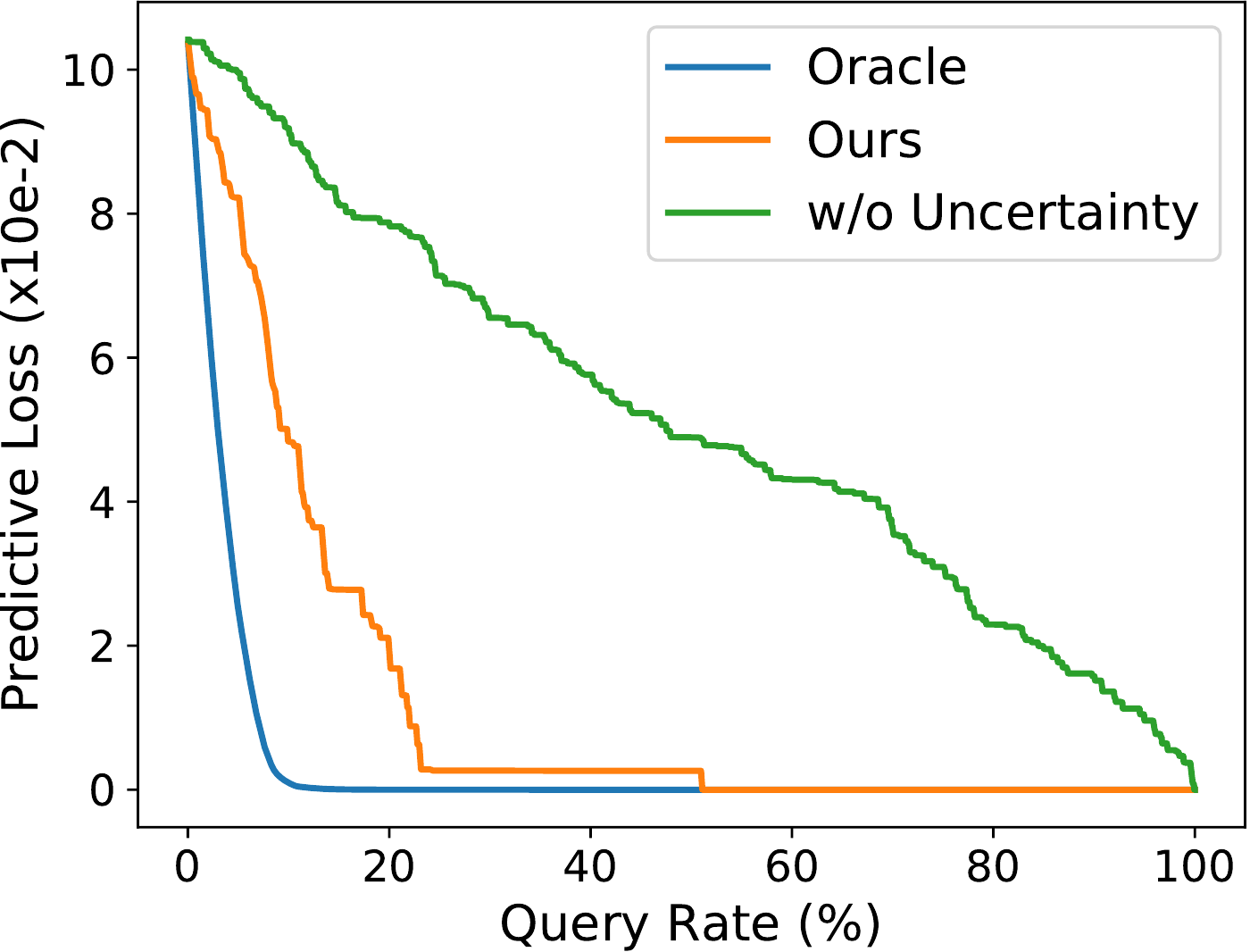}
\end{minipage}
}%
\subfigure[AUC-ROC]{
\begin{minipage}[t]{0.15\textwidth}
\centering
\includegraphics[width=.985\textwidth]{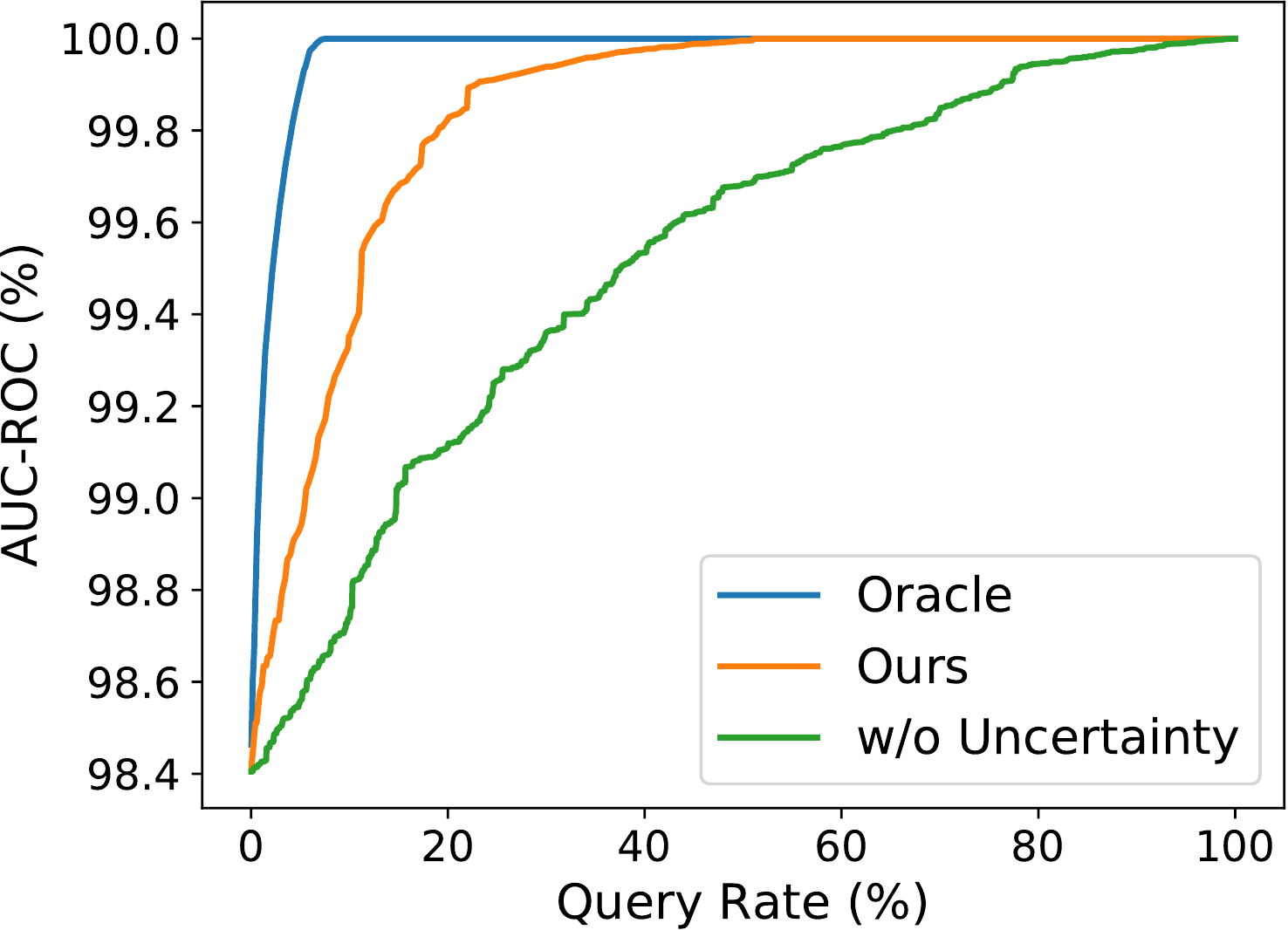}
\end{minipage}%
}%
\subfigure[Average Precision (AUC-PR)]{
\begin{minipage}[t]{0.15\textwidth}
\centering
\includegraphics[width=.985\textwidth]{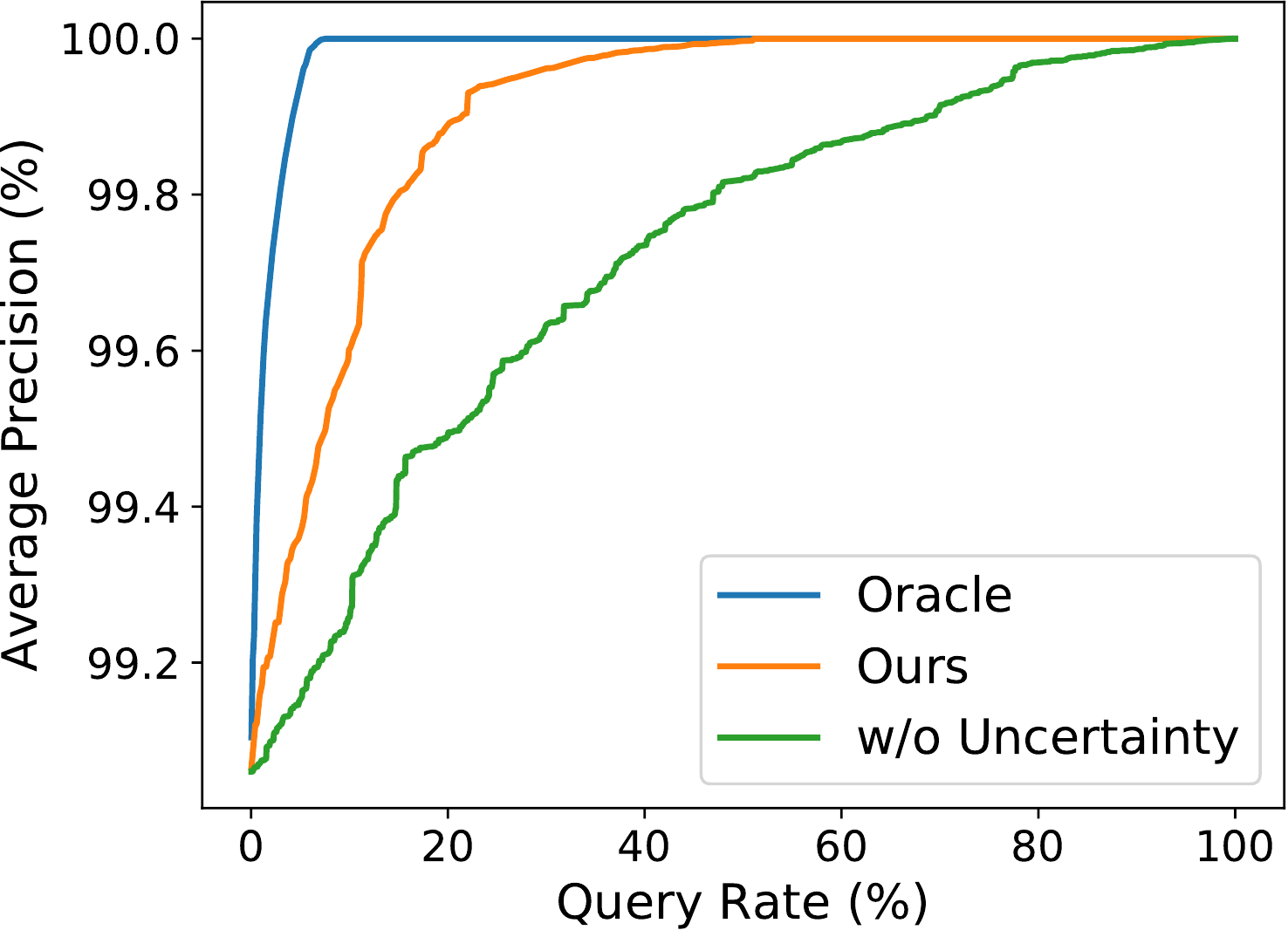}
\end{minipage}
}%
\centering
\caption{\textit{Performance change tendency at different query rates.} Note: around 20 \% query rate, the model accuracy has become significantly better - nearly perfect. Query rate: it is defined as the lower confidence prediction of sample versus the total of test sample. }\label{fig:overall}
\end{figure}
\paragraph{Results}As depicted in Figure~\ref{fig:overall}, we make comparisons with three metrics at various query rates. It is observed that our model quickly reduces the predictive bias ($\mathscr{l_2}$ testing loss) on test data from Figure~\ref{fig:bias}. Its predictive bias decreases to nearly $0$ with only about $20\%$ query samples, whereas the uncertainty-agnostic model requires almost $100\%$ query data to achieve similar performance. In terms of the remaining three measurements, our approach also outperforms the uncertainty-agnostic one by a large margin. For quantitative evaluation, we report the performance gain w.r.t AUC-ROC and AUC-PR in Table~\ref{tab:roc} and Table~\ref{tab:pr} at various query rates. The purpose of this table is to show with such queried data, the labels of the data will be corrected correspondingly thus leading to the increase of performance gain. The larger increase of the performance gain indicates the increased likelihood of the model to predicate the uncertain sample. Along with the growth of query rates, the performance gain of our model rises much faster than uncertainty-agnostic predictions.
\begin{table}[h!]\small
  \caption{\emph{Performance gain (\%) of AUC-ROC at various query rates.} The value of AUC-ROC at $0\%$ query rate is $98.40\%$.}
  \label{tab:roc}
  \centering
  
\begin{tabular}{cllllllll}
\hline
\textbf{Methods} & \multicolumn{1}{c}{\textbf{0.1\%}} & \multicolumn{1}{c}{\textbf{0.5\%}} & \multicolumn{1}{c}{\textbf{1\%}} & \multicolumn{1}{c}{\textbf{5\%}} & \multicolumn{1}{c}{\textbf{10\%}} & \multicolumn{1}{c}{\textbf{50\%}} \\ \hline
Oracle           & 0.18                               & 0.47                               & 0.73                             & 1.50                             & 1.60                     & 1.60                              \\
w/o Uncertainty           & 0.00                               & 0.01                               & 0.02                             & 0.15                             & 0.33                     & 1.28                              \\
Ours             & 0.03                               & 0.11                               & 0.18                             & 0.54                             & 0.95                     & 1.60                              \\ \hline
\end{tabular}
\end{table}

\begin{table}[h!]\small
  \caption{\emph{Performance gain (\%) of average precision (AUC-PR) at various query rates.}  The value of average precision at $0\%$ query rate is $99.06\%$.}
  \label{tab:pr}
  \centering

\begin{tabular}{cllllllll}
\hline
\textbf{Methods} & \multicolumn{1}{c}{\textbf{0.1\%}} & \multicolumn{1}{c}{\textbf{0.5\%}} & \multicolumn{1}{c}{\textbf{1\%}} & \multicolumn{1}{c}{\textbf{5\%}} & \multicolumn{1}{c}{\bf 10\%} & \multicolumn{1}{c}{\textbf{50\%}} \\ \hline
Oracle           & 0.13                               & 0.31                               & 0.47                             & 0.89                             & 0.94                     & 0.94                              \\
w/o Uncertainty           & 0.00                               & 0.01                               & 0.01                             & 0.09                             & 0.20                     & 0.76                              \\
Ours             & 0.02                               & 0.06                               & 0.11                             & 0.31                             & 0.54                     & 0.94                              \\ \hline

\end{tabular}
\end{table}

\subsection{Exploration and Ablation Studies}\label{sec:ex}
\begin{figure}[h!]
\centering
\subfigure[Legend]{
\begin{minipage}[t]{0.15\textwidth}
\centering
\includegraphics[width=.65\textwidth]{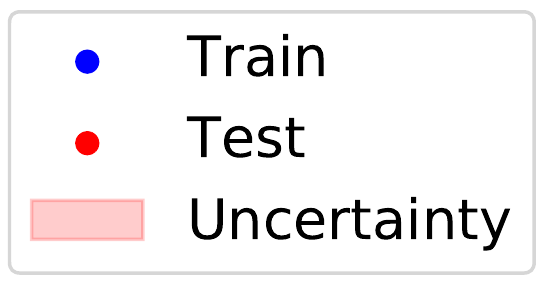}
\end{minipage}
}%
\subfigure[Worst Radius]{\label{fig:worstR}
\begin{minipage}[t]{0.15\textwidth}
\centering
\includegraphics[width=.985\textwidth]{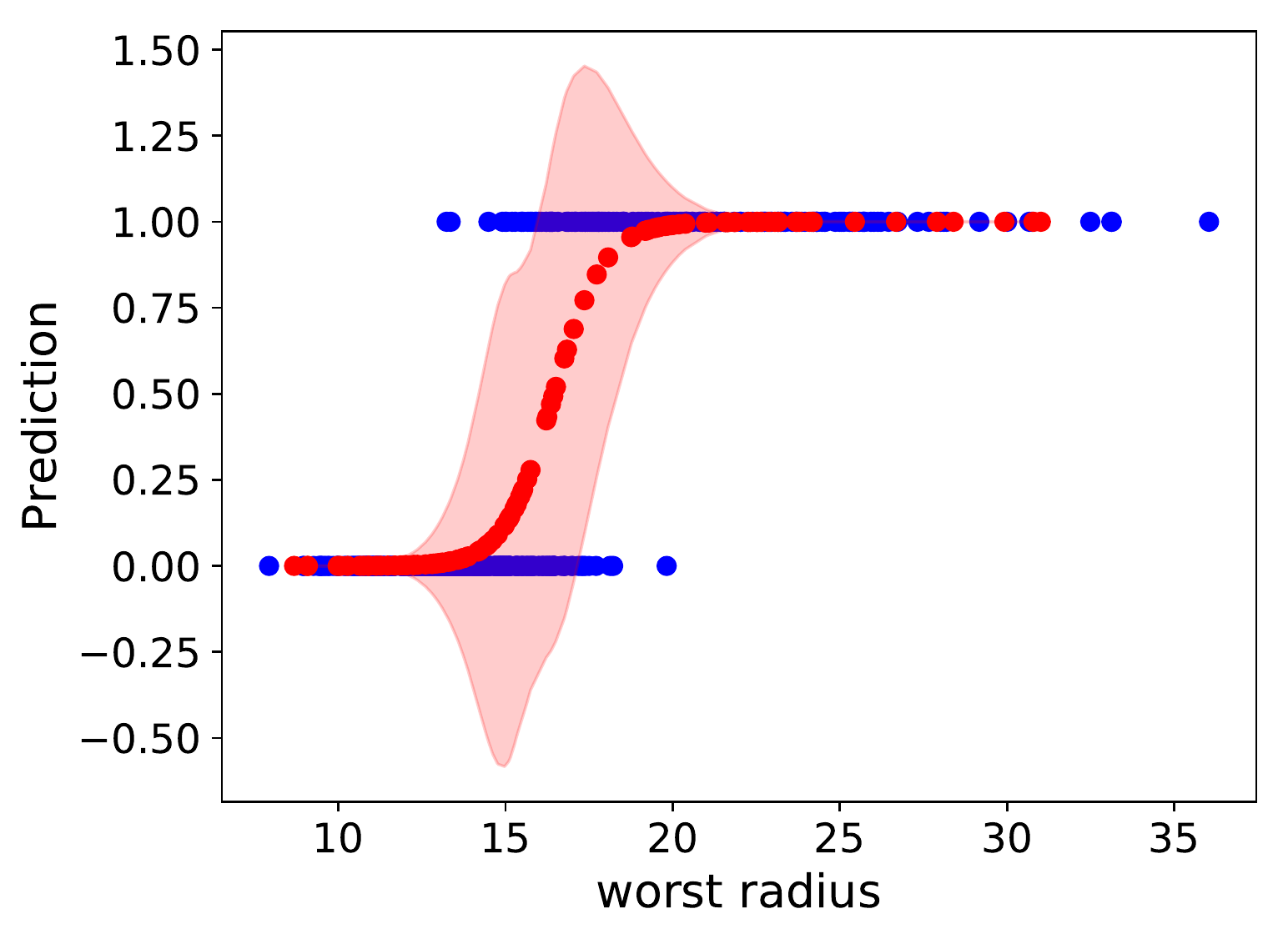}
\end{minipage}
}%
\subfigure[Perimeter Error]{
\begin{minipage}[t]{0.15\textwidth}
\centering
\includegraphics[width=.985\textwidth]{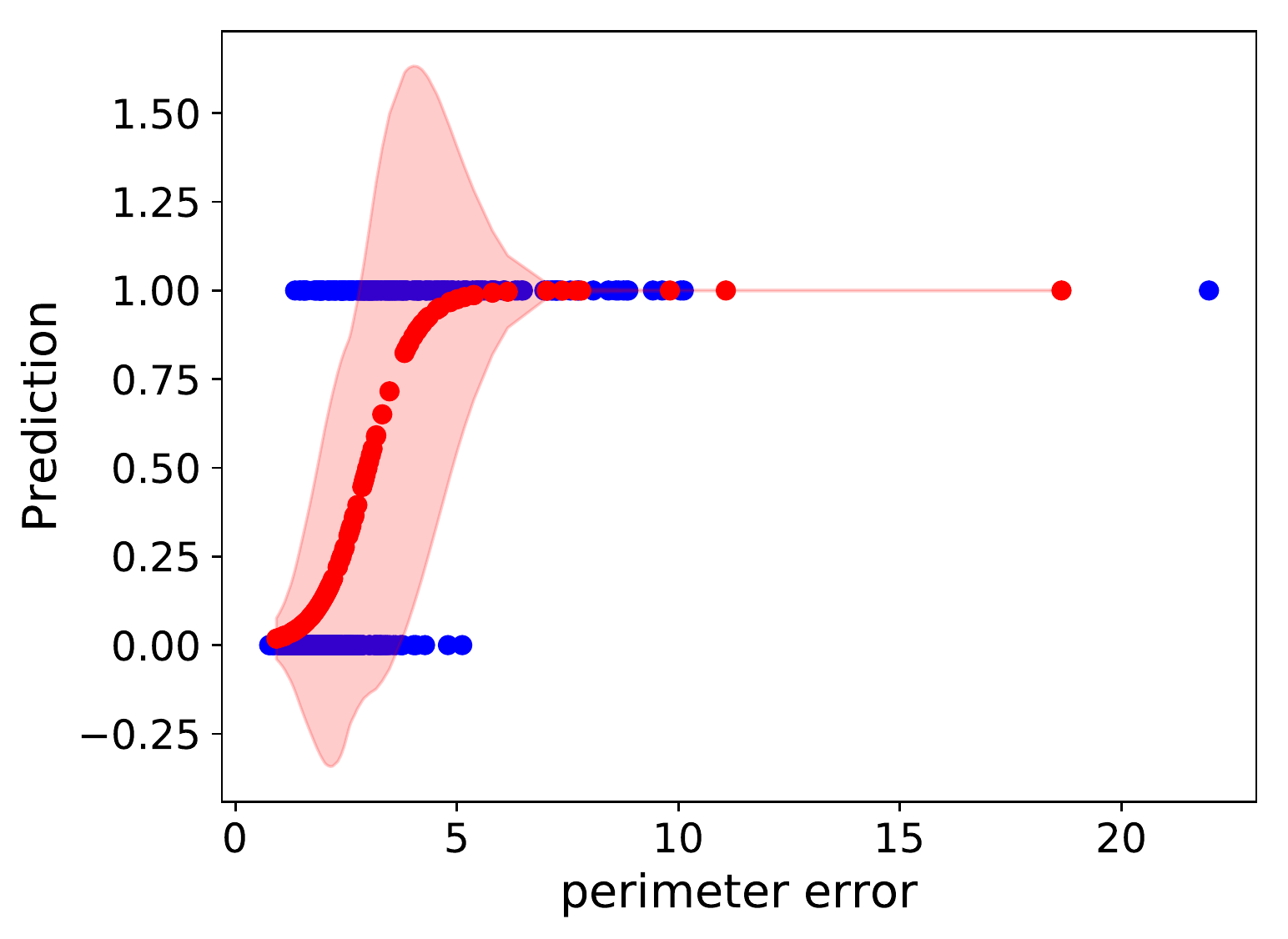}
\end{minipage}
}%

\subfigure[Mean Area]{
\begin{minipage}[t]{0.15\textwidth}
\centering
\includegraphics[width=.985\textwidth]{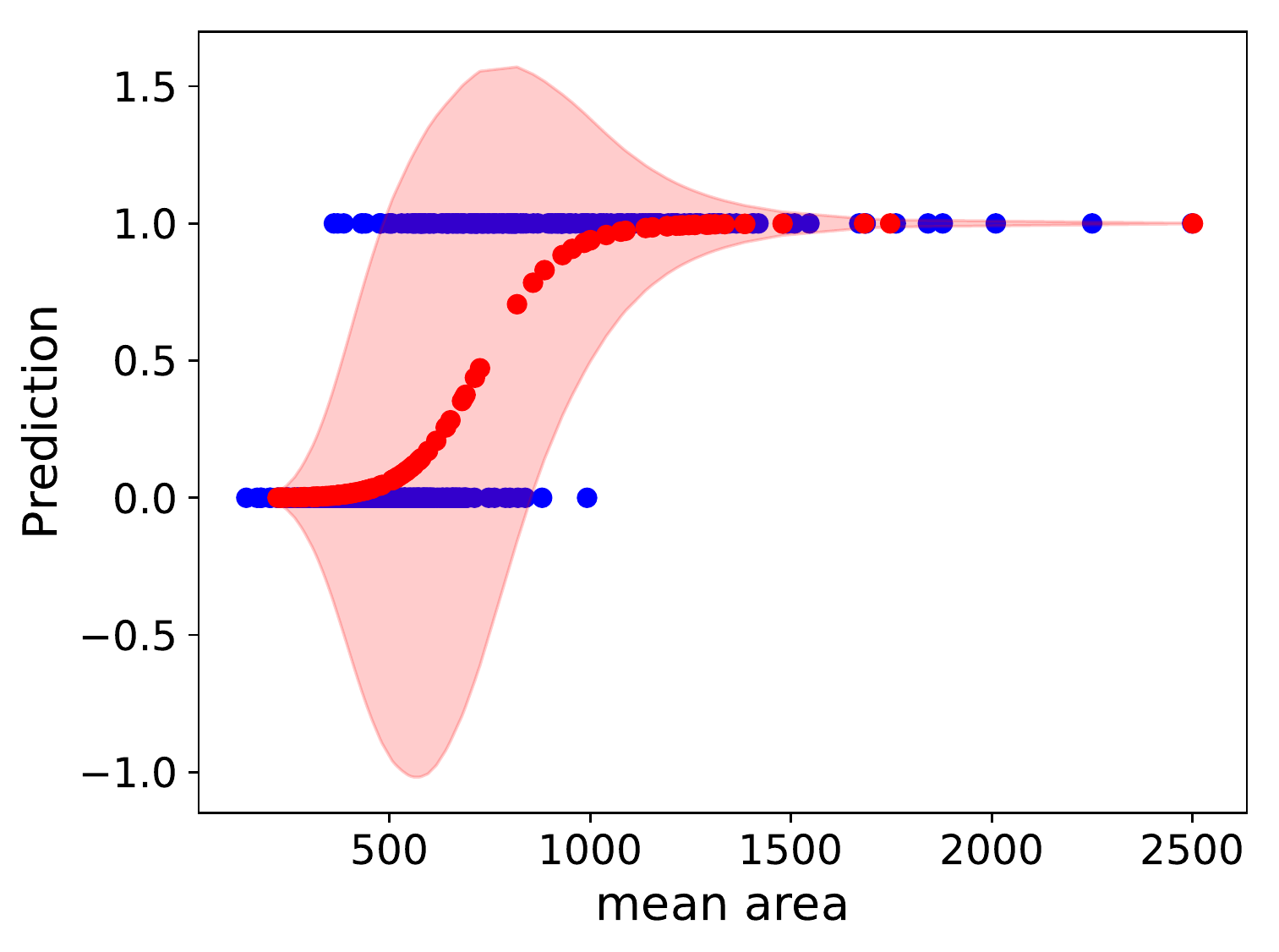}
\end{minipage}
}%
\subfigure[Smoothness Error]{
\begin{minipage}[t]{0.15\textwidth}
\centering
\includegraphics[width=.985\textwidth]{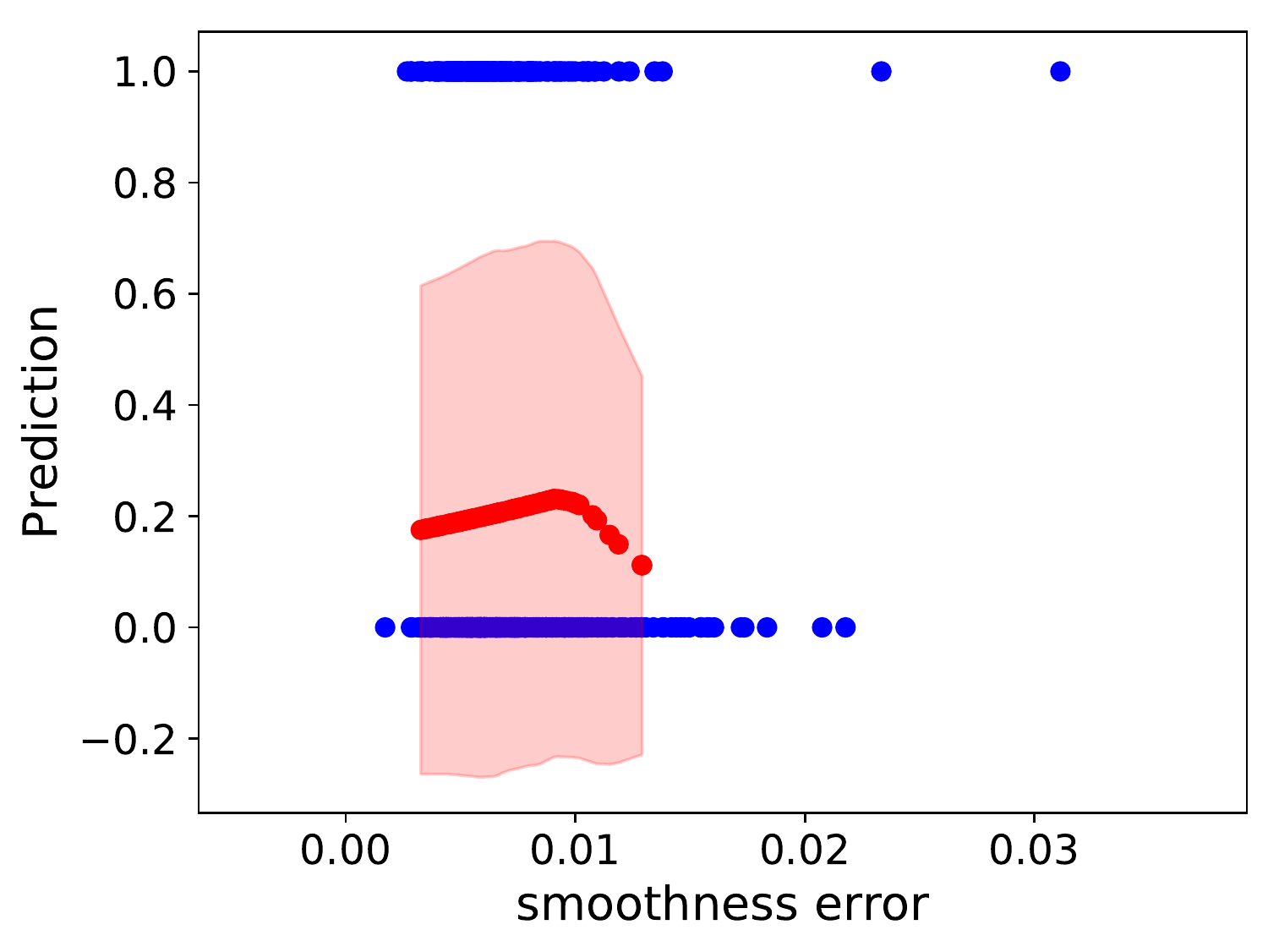}
\end{minipage}
}%
\subfigure[Worst Concavity]{\label{wc}
\begin{minipage}[t]{0.15\textwidth}
\centering
\includegraphics[width=.985\textwidth]{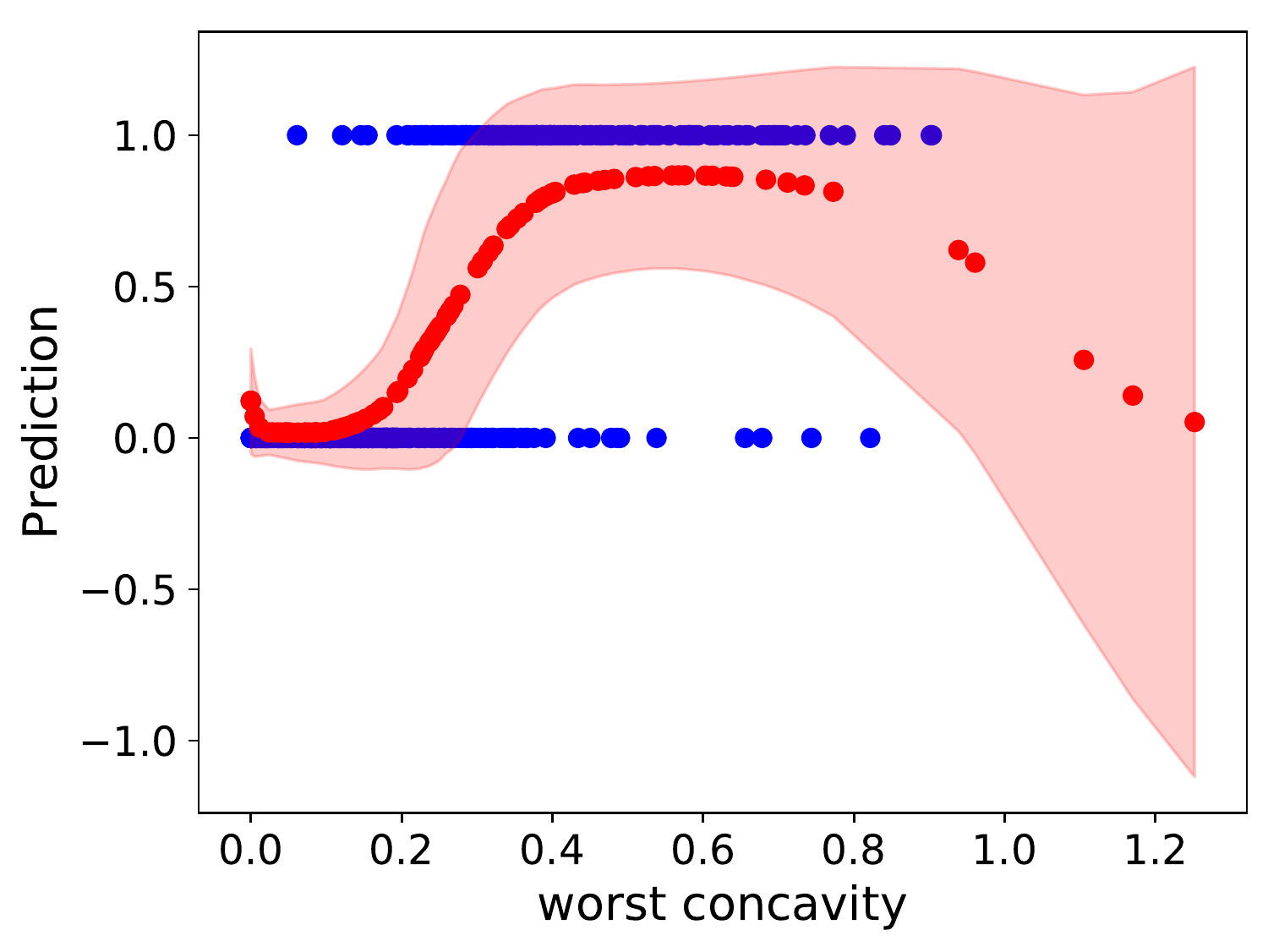}
\end{minipage}
}%
\centering
\caption{\textit{Exploration of uncertainty quantification.} The x-axis is the value of a certain feature and the y-axis represents the testing prediction (the training label). The red shade is the area with $x=$\textit{feature value} and $\mu({\bf x},{\bf s};\psi)- \sigma({\bf x},{\bf s};\psi) \leq y \leq \mu({\bf x},{\bf s};\psi)+ \sigma({\bf x},{\bf s};\psi)$. }\label{fig:ab}
\end{figure}

In this paper, two additional modules is introduced to enhance the vanilla INVASE, \textit{i.e.}, uncertainty quantification of the predictor and reward shaping of the selector. Through exploration and ablation studies, we attempt to verify their efficacy individually. It is important to note that the shaded area in Figure~\ref{fig:ab} represents the uncertainty of the model to predicate the correct positive or negative sample.    

\paragraph{Can the uncertainty quantification module capture confidence?} As shown in Figure~\ref{fig:worstR} w.r.t. the feature ``Worst Radius'', the test of uncertainty within the range from 12 to 20 is large since the training labels are ambiguous (mix of 0 and 1), whereas the uncertainty beyond that range is approximately 0 because the training annotations are exclusively 0 or 1. Similar correspondences also occur in the remaining features of Figure~\ref{fig:ab}.  The aforementioned results show that our uncertainty quantification module is helpful to uncertainty modeling. 

\begin{figure}[h!]
\centering
\subfigure[w/ Reward Shaping]{\label{fig:o1}
\begin{minipage}[t]{0.22\textwidth}
\centering
\includegraphics[width=.985\textwidth]{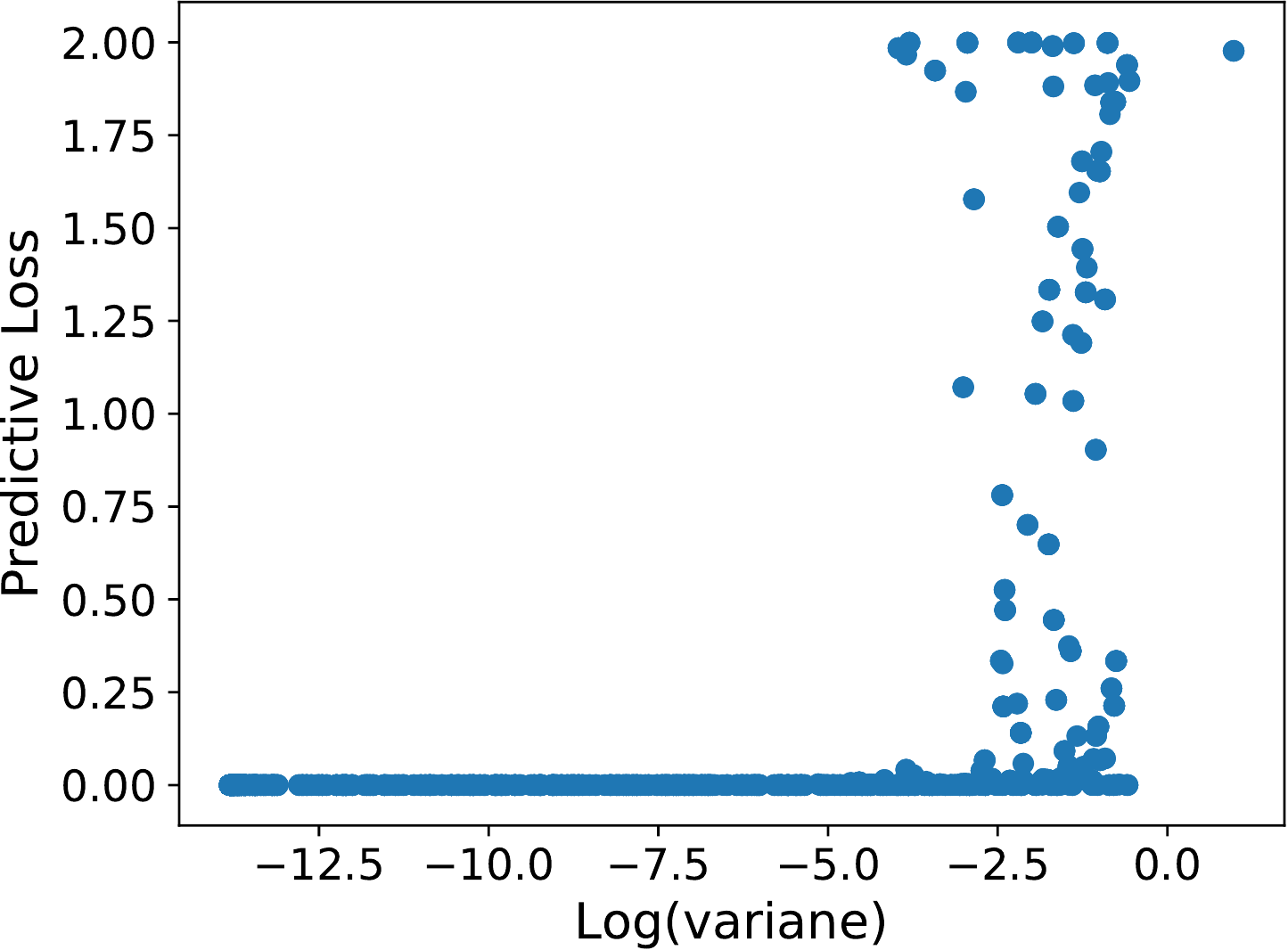}
\end{minipage}
}%
\subfigure[w/o Reward Shaping]{\label{fig:o0}
\begin{minipage}[t]{0.22\textwidth}
\centering
\includegraphics[width=.985\textwidth]{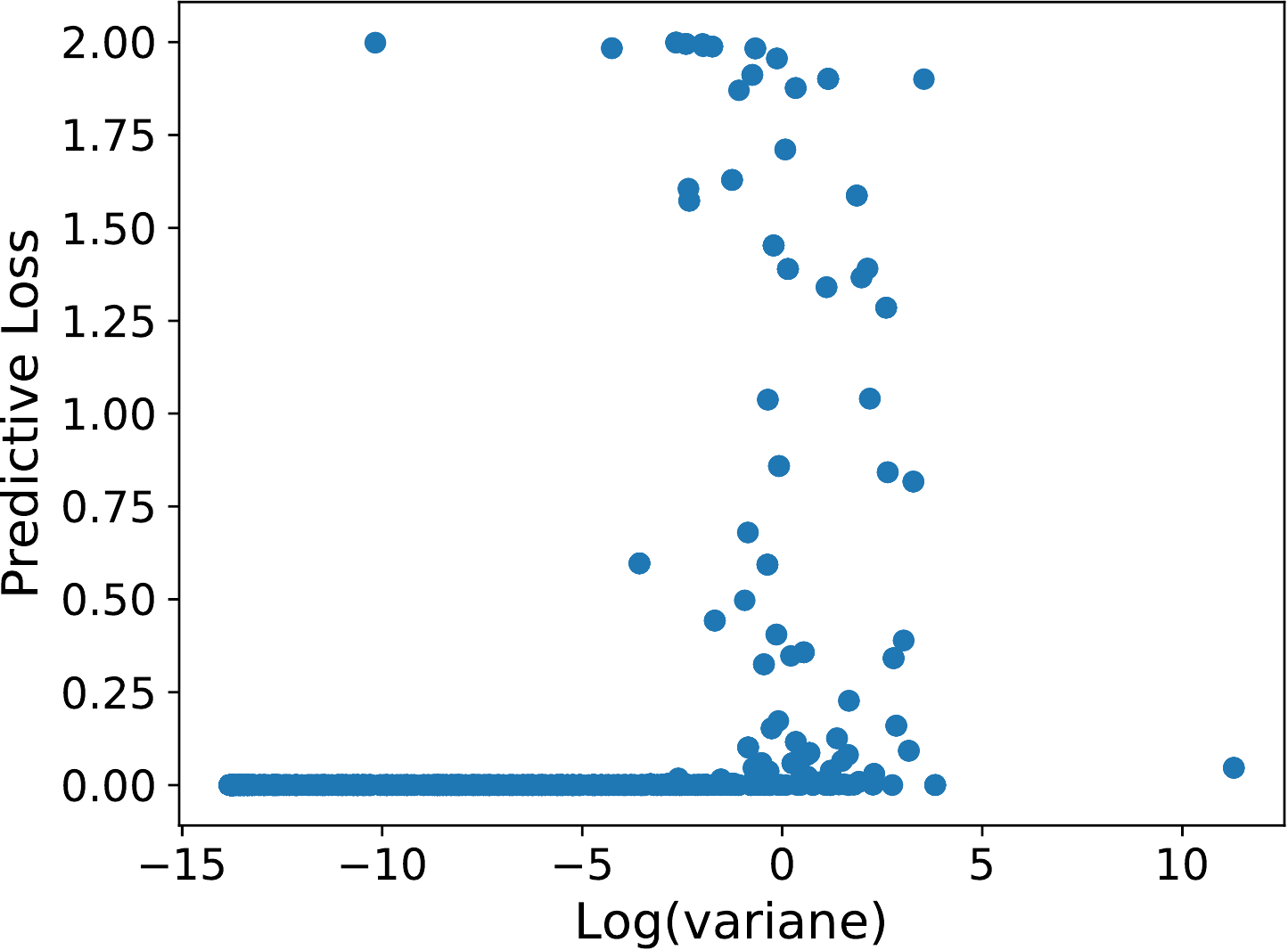}
\end{minipage}
}%
\centering
\caption{\textit{Ablation of reward shaping.} (Left: with reward shaping, Right: without reward shaping) The x-axis is the value of $\log\sigma^2$, the log of predicative variance (Equation~(\ref{log_l})) and the y-axis represents the predictive bias for the corresponding test data points.}\label{fig:omega}
\end{figure}
\paragraph{Can the reward shaping module benefit uncertainty estimation?} We adopt the predictive bias to investigate whether uncertainty is correctly estimated, As shown in Figure~\ref{fig:o1}, variances learned with reward shaping are indeed effective indicators for mis-classified results. However, variances of the model without reward shaping scatter across a wider range of the x-axis and have weaker relevance with the predictive bias. 

\section{Conclusion}
In this paper, we present an uncertainty-aware INVASE to quantify predictive confidence. The model is able to quantify the potential errors of our instance-wise feature selection, which may be beneficial to some healthcare problems. In theory, the proposed approach extends the modeling perspective from a Dirac delta function to a learnable uncertainty-aware distribution. Conceptually, it is a highly scalable framework, of which any distribution with differentiable parameters is an eligible tool. To be specific, we apply Gaussian distributions to capture uncertainty with their variances. Accordingly, we implement the uncertainty-aware model with two extra modules over the raw INVASE, \textit{i.e.}, uncertainty quantification of the predictor and reward shaping of the selector. To evaluate our method, we carry out experiments on UCI-WDBC w.r.t. the whole model and each new component. Experimental results show that our approach discovers overwhelming majority testing errors with only about $20\%$ queries, whereas the uncertainty-agnostic counterparts need nearly $100\%$ query samples for the same performance gain.   
\bibliography{sample}

\end{document}